\documentclass[11pt]{article}

\usepackage{times}
\usepackage{graphicx}
\usepackage{amsmath}
\usepackage{amssymb}
\usepackage{hyperref}
\usepackage{geometry}
\usepackage{algorithm}
\usepackage{algorithmic}
\usepackage{caption}
\usepackage{placeins}
\usepackage{bm}

\title{Mitigating Catastrophic Forgetting in the Incremental Learning of Medical Images}

\author{
Sara Yavari, Jacob Furst\\[0.5em]
College of Computing and Digital Media, DePaul University, Chicago, IL, USA\\[0.5em]
\texttt{syavari@depaul.edu}, \texttt{jfurst@cdm.depaul.edu}
}

\date{}
\begin{document}

\maketitle

\begin{abstract}
This paper proposes an Incremental Learning (IL) approach to enhance the accuracy and efficiency of deep learning models in analyzing T2-weighted (T2w) MRI medical images prostate cancer detection using the PI-CAI dataset. We used multiple health centers’ artificial intelligence and radiology data, focused on different tasks that looked at prostate cancer detection using MRI (PI-CAI). We utilized Knowledge Distillation (KD), as it employs generated images from past tasks to guide the training of models for subsequent tasks. The approach yielded improved performance and faster convergence of the models. To demonstrate the versatility and robustness of our approach, we evaluated it on the PI-CAI dataset, a diverse set of medical imaging modalities including OCT and PathMNIST, and the benchmark continual learning dataset CIFAR-10. Our results indicate that KD can be a promising technique for IL in medical image analysis in which data is sourced from individual health centers and the storage of large datasets is not feasible. By using generated images from prior tasks, our method enables the model to retain and apply previously acquired knowledge without direct access to the original data.
\end{abstract}

\textbf{Keywords:} Incremental Learning, Medical Images, Knowledge Distillation.

\section{Introduction}

Incremental Learning (IL) is a Deep Learning paradigm in which the model continuously ‘learns’ from sequential input data without access to previous data \cite{Re43}. A finite sequence of tasks (experiences) is provided to the model across the timeline with the goal of remembering all prior tasks when starting each new task. In the IL paradigm, the concept of ‘task’ (or ‘experience’) refers to a distinct training phase with each new data set, that is connected to a unique experience, a new domain, or a diverse output space \cite{Re1}. In practice, however, IL suffers from the critical issue of catastrophic forgetting, in which the model substantially or entirely ‘forgets’ what it has learned \cite{Re1, Re2}. Catastrophic forgetting presents a significant challenge in machine learning’s practical applications in dynamic continual learning environments such as healthcare \cite{Re3}, \cite{Re5}, \cite{ReNing}. Catastrophic forgetting can lead to decreased accuracy as models struggle to integrate new insights with existing knowledge. While IL has demonstrated effectiveness in medical image processing to assist physicians, it faces challenges due to extensive data requirements, patient privacy concerns, and storage limitations \cite{Re44}. 

An emerging technique in tackling the challenges of IL is Knowledge Distillation (KD) \cite{re103}. The technique is well suited at mitigating catastrophic forgetting as it transfers learned representations from the larger and more complex model (the Teacher) to the more efficient and smaller model (the Student) with the introduction of new tasks \cite{re101}. Using the KD technique, the Student model can retain knowledge from prior tasks, and adapt to new information as it becomes available. The Teacher model is used by KD to allow the Student model to capture the nuanced relationships between classes, thereby preserving the decision boundaries that were learned from prior tasks\cite{re100}, \cite{re102}. In the analysis of medical images, maintaining integrity of acquired prior knowledge is critical for diagnoses that are accurate, and treatment plans that are appropriate. Thus, KD can preserve model performance, maintain anonymity, and reduce data storage requirements \cite{re103}, which often represent real-world challenges in healthcare.

Our motivation is employing an IL method with KD for medical images that can effectively use data from multiple medical centers and prevent catastrophic forgetting. In lieu of retaining actual samples from previous tasks, we generate synthetic samples using a previously trained (fixed) model. The approach allows AI models to consistently learn new information while retaining the medical knowledge previously obtained. Our approach utilizes a shallow generator to produce a subset of historical data, and introduces a unique KD loss to combat catastrophic forgetting effectively. The key contributions of the method explored include: A novel Incremental Learning (IL) method that utilizes metric KD, incorporating feature attention matching loss and a covariance matrix within the embedding space (a continuous vector space capturing data features) to reduce the forgetting of previous knowledge; and a technique for synthesizing past data samples using a shallow Variational Autoencoder (VAE), where the teacher model guides the generation process to adhere closely to the gold standard distribution of past data.

\section{Related Work}
Incremental Learning (IL) is alternatively referred to as Continual Learning (CL) \cite{ref107} or lifelong learning \cite{ref108}. IL has emerged as a valuable technique for enhancing the performance and adaptability of deep learning models in medical image analysis. CL \cite{ref109,ref110,ref111} involves the challenge of sequentially learning multiple tasks while retaining and leveraging previously acquired knowledge to enhance performance on subsequent tasks. To address the challenges associated with CL, including catastrophic forgetting, a few approaches have been devised \cite{ref112}. These approaches can be classified into three main categories: Rehearsal-based methods, Regularization-based methods, and Architecture-based methods \cite{ref112, ref113}. Rehearsal-based methods \cite{ref114, ref115,ref116, ref117} store condensed data summaries from previous tasks, replaying them while training, and maintaining obtained knowledge. This approach helps mitigate the issue of catastrophic forgetting by periodically revisiting and reinforcing past learning experiences. Some techniques within this approach \cite{ref118,ref119} employ generative models, for instance Variational Autoencoders (VAEs) or Generative Adversarial Networks (GANs), to create synthetic or pseudo rehearsal data that resembles samples from earlier tasks.
In study \cite{ref120}, a rehearsal-based CL with active learning was integrated to choose the most instructive samples. Although rehearsal methods are favored for their high performance, their dependency on the availability of past data makes them less suitable for medical applications due to patient data privacy concerns \cite{ref113}. Rehearsal-based continual learning necessitates extra storage for maintaining the replay buffer or generative models \cite{ref112}. By setting constraints and limiting modifications to weights or nodes associated with previously learned tasks, regularization-based methods \cite{ref121,ref122,ref123,ref124} help mitigate catastrophic forgetting \cite{ref125}. These methods often incorporate knowledge distillation \cite{ref113} to enhance performance by transferring insights from a larger to a smaller model, preserving and using past knowledge effectively. They also use KD in CL, transferring knowledge from a teacher to a student, or a larger and more complex model to a smaller and simpler model respectively. This technique is pivotal for enhancing generalization, increasing accuracy, and mitigating catastrophic forgetting \cite{Re1}. Regularization-based methods enforce constraints on model parameters (e.g., \cite{ref126,ref127,ref128}) to maintain knowledge acquired from prior tasks. In \cite{ref129} proposed transferring knowledge using not only the logit layer but also earlier layers. To address differences in layer width, they introduced a regressor to connect the intermediate layers of the teacher and student models. However, a standardized approach to achieve this is still lacking \cite{ref130}. To address this issue, \cite{ref131, ref132} employed a shared representation of layers. Nevertheless, selecting the appropriate layer for matching remains a complex challenge. In \cite{ref133}, the authors increased small network accuracy, and preserved features and knowledge that were critical in large networks. In the distillation process of \cite{ref134}, student and teacher scores were artificially assigned, which may potentially lead students to learn less relevant information. The authors of \cite{ref135} introduced a feature-level KD technique employing contrastive learning for continuous nuclei segmentation. The authors of \cite{ref136} proposed distillation in the mixed-curvature space of embedding vectors to preserve the intricate geometric structure of medical data. These methods face a significant challenge in striking a balance between effectively learning new tasks and maintaining the valuable knowledge gained from previous tasks \cite{ref134}. This is essential for ensuring sustained adaptation and continual enhancement of performance over time. Achieving this balance allows models to effectively integrate new knowledge while retaining the valuable insights gained from prior experiences \cite{ref137}. Architecture-based methods,\cite{ref138,ref139, ref140}. Utilize the capacity to either extend \cite{ref141}, or segregate \cite{ref142} model parameters, aiming to preserve acquired knowledge and prevent forgetting. While effective in mitigating catastrophic forgetting, these methods often require additional capacity for task-adaptive parameters \cite{ref138} or replay buffers. Our proposed framework combines knowledge distillation with regularization strategies and employs a VAE to generate rehearsal samples that resemble the previous task’s data distribution. This approach avoids retaining actual samples (due to privacy concerns) and eliminates the need for storing data, reducing storage requirements and computational complexity.

\section{Method}
\subsection{Preliminaries}
% This section provides an overview of the main components that form the basis of our proposed method, including incremental learning, metric learning techniques and Discriminative-separable Feature Space and Task Confusion.
\subsubsection{Incremental Learning Formulation}
A model is initially trained from scratch using the complete training dataset $D_1$, which contains $C_1$ classes. Subsequently, new datasets $D_k$, each containing $C_k$ classes, are gathered sequentially. Each dataset $D_k$ is used to train a model $M_k$ incrementally. Specifically, model $M_k$ is trained only on dataset $D_k$ for each incremental task $k = 2, \ldots, K$, where $K$ denotes the total number of incremental tasks. The current model $M_k$ doesn't have access to previous real data, meaning that the problem is zero-shot class-incremental learning. The parameters of the deep neural network model trained on $D_k$ are denoted as $\bm{\theta_{k}}$.

\subsubsection{Metric Learning and Task Confusion}
The goal of deep metric learning is to train a differentiable network (feature extractor) $f_\theta (.)$ ($\theta$ are learned weights) that maps an input domain $X$ into an embedding compressed domain$f_\theta (.): X\rightarrow\mathbb{R}^D$ together with a distance metric $D:\mathbb{R}^D\rightarrow\mathbb{R}$ in a way such that two similar data samples result in a small distance value, and dissimilar data points produce a large distance value. Various loss functions are proposed for learning similarities, such as contrastive loss \cite{Re31}, center loss \cite{Re32}, and triplet loss \cite{Re33}, the latter of which we use in our IL proposed method. The triplet loss uses three instances: an anchor $(x_a)$, a positive instance with the same identity as the anchor $(x_p)$, and a negative instance with a different identity $(x_n)$. Assuming $D(i,j)$ is a metric on the embedding space (e.g., squared Euclidean distance), the triplet loss is defined as $\mathcal{L}_{\text{tri}}$ for a single triplet input $(x_a,x_p,x_n)$. For simplicity, we use $(a,p,n)$ as the triplet, $\mathcal{L}_{\text{tri}} = \max(0, D(a,p) - D(a,n) + m) \label{eq3}$. The purpose of triplet loss is satisfying $D(a, p) - D(a, n) + m \leq 0$, which decreases the distance between the positive samples and the anchor, $D(a, p)$, in comparison to the negative samples and the anchor, $D(a, n)$, by predefined margins $m$. Using deep metric learning as the backbone, we use the Nearest Class Mean classifier during test time: $y^* = \arg\min_{v \in S} \text{dist}(f_\theta(x), \mu_v)$. Where $dist(i, j)$ is a distance metric (e.g., Euclidean distance as in our simulation) and $\mathcal{S}$ denotes the set of all class indices in the test data. In this equation, the embedding centroid $\mu \in \mathbb{R}^p$ of class $v$ is given by: $\mu_v=\frac{1}{n_v}\sum_{i=1}^{J}f_\theta(x_i)[y_i=v],\label{eq3}$ where $J$ is total number of data points in input domain $\mathcal{X}$, $n_v$ denotes the number of data points in class $v$, $[Q] = 1$ if condition $Q$ is true; otherwise $[Q] = 0$. In the course of IL, for each class, $\mu_v$ is computed and stored, which is used later for inference. 

% \begin{figure}[!t]
% \captionsetup[figure]{font=scriptsize}
% % Caption and label go in the first argument and the figure contents go in the second argument
% \floatconts
%   {fig2}%
%   {\caption{a) Separable features b) Discriminative features.}}%
%   {\includegraphics[width=0.47\linewidth]{Fig3.jpg}} % Adjusted to match your original scale
% \end{figure}
\begin{figure}[htbp]
\centering
\includegraphics[width=0.47\linewidth]{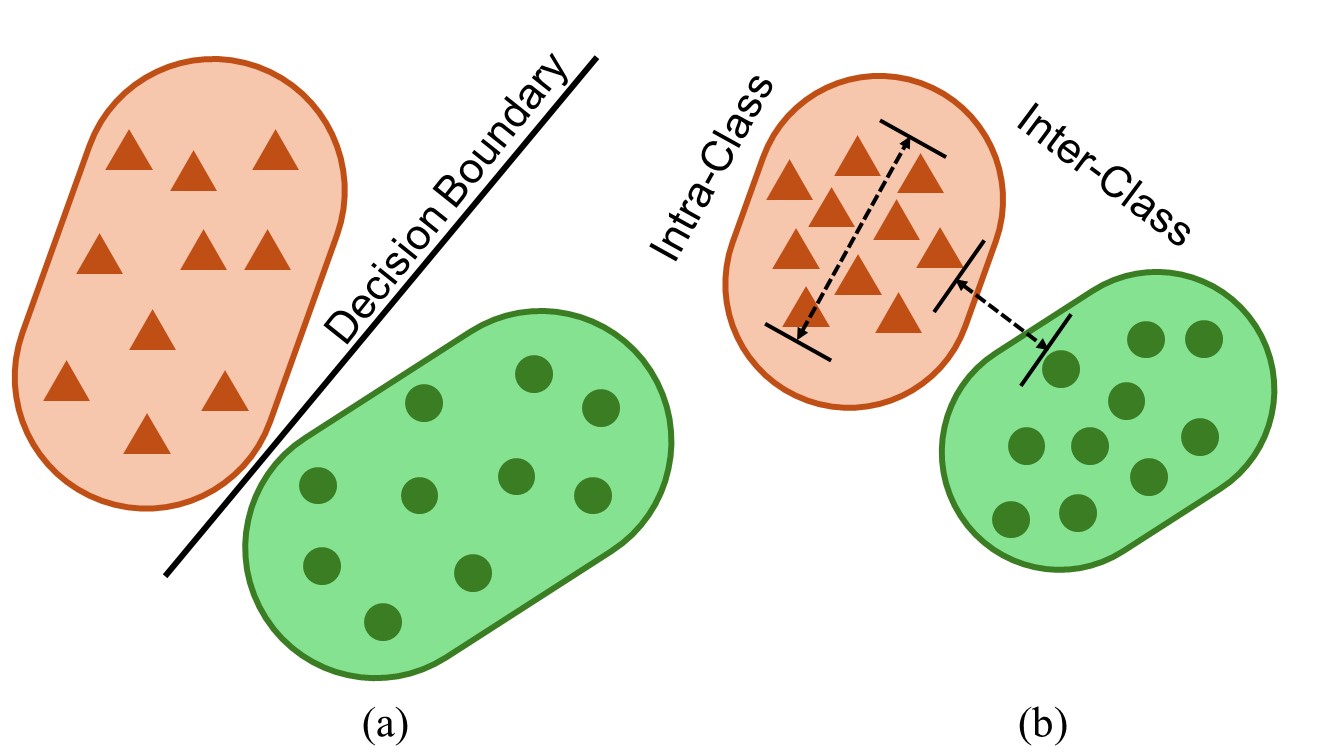}
\caption{a) Separable features. b) Discriminative features.}
\label{fig:fig2}
\end{figure}

Task confusion is a key issue in continual learning which is further elaborated on in Appendix A. Softmax causes task confusion with classes in new and previous tasks by affecting the inter- and intra-class distances, \cite{Re99}, Figure~\ref{fig:fig2}(a). A solution to task confusion is to make classes in the embedding space both separable and discriminative, as shown in Figure~\ref{fig:fig2}(b). While generative models can achieve this by learning each class separately \cite{Re34}, \cite{Re35}, scalability remains a challenge. 

\begin{figure}[t]
% \captionsetup{aboveskip=0pt, belowskip=0pt}
% \setlength{\abovecaptionskip}{27pt} % Remove space above the caption
% \setlength{\belowcaptionskip}{27pt} % Remove space below the caption
\centering
\includegraphics[width=0.82\linewidth]{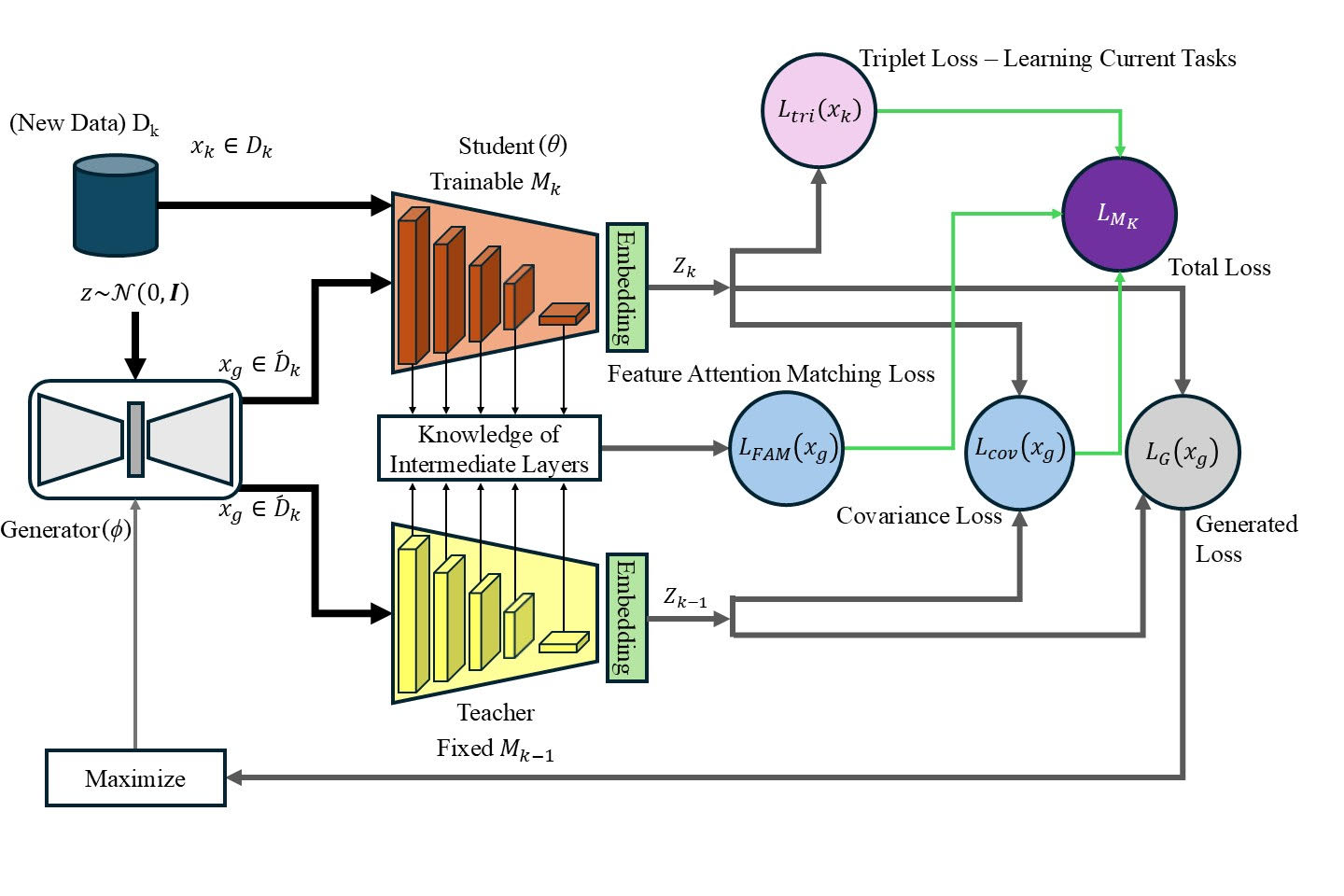}
\caption{Overview diagram of the proposed framework: The triplet loss \( \mathcal{L}_{\text{tri}}(x_k) \) is used to train the Student model \( M_k \) on the current task using real data (\( x_k \in D_k \)). Distillation (KD) loss, defined as \( \mathcal{L}_{\text{KD}} = \mathcal{L}_{\text{FAM}} + \mathcal{L}_{\text{cov}} \), mitigates catastrophic forgetting. A light Variational Autoencoder (VAE) serves as a generator, with the loss (\( \mathcal{L}_{G} \)) maximizing the distance between the Student (\( M_k \)) and Teacher (\( M_{k-1} \)) adversarially, ensuring synthetic images resemble previous task data distribution and enabling effective knowledge transfer.}
\label{fig:proposed-method}
\end{figure}

\subsection{Proposed Framework}
Figure~\ref{fig:proposed-method} illustrates the overview of our method. In each task, new data $x_k \in D_k$ will go through the Student model $M_k$, and triplets are generated online to learn the current task $k$. We generate synthetic data from past tasks $x_g \in \hat{D}_k$ to combat catastrophic forgetting using our proposed KD loss. To generate synthetic past data $\hat{D}_k$, we use adversarial training between Student and Teacher inspired by~\cite{Re89} (more details in Subsection~\ref{generat}). Total losses for training the Student $M_k$ includes three terms as:
\begin{equation}
\mathcal{L}_{M_k} = \mathcal{L}_{\text{tri}}(x_k) + \lambda \mathcal{L}_{\text{KD}}(x_g) + D_{E}(M_{k}, M_{k-1}), \label{eq14}
\end{equation}

where $\mathcal{L}_{\text{tri}}$ is the triplet loss for leaning current task $k$ from current dataset $D_k$. The $\mathcal{L}_{\text{KD}}$ loss combats catastrophic forgetting by transferring knowledge from the fixed Teacher to the Student (Subsection~\ref{KDloss}). To achieve this, the term $D_E(\cdot)$ measures the distance between the Teacher and Student representations when processing synthetic data generated by the VAE. Additionally, $D_E(\cdot)$ ensures that the generated data preserves the distribution of past tasks as learned from the Teacher model (Subsection~\ref{generat}).

\subsubsection{Generating $\hat{D}_k$} \label{generat}
To train the model using medical images from a new medical center (current task $k$), and not forget information provided by previous medical centers, we generate synthetic images for the previous task. We use the generated images to distill knowledge from the last task into the current model. This is to store and re-use a previously trained model for generating the synthetic images used for training the current model. To generate synthetic images $x_g \in \hat{D}_k$ we employ a shallow VAE with parameters $\phi$ that is trained adversarially with the (fixed) $M_{k-1}$ as the Teacher and $M_k$ as the Student. Our approach avoids using generated samples alongside current real samples. Our model knows previously learned distributions by $M_{k-1}$ and uses them to mitigate catastrophic forgetting as it is learning the current task $k$. The generator's objective is to maximize the distance between the Teacher $M_{k-1}$ and the Student $M_k$ in an adversarial manner, compelling it to adhere closely to the distribution learned by the Teacher by minimizing $\mathcal{L}_{G}$ in $n_g$ iterations as follows:
\begin{equation}
\mathcal{L}_{G} = -D_{E}\bigl(M_{k}(\boldsymbol{x_g}), M_{k-1}(\boldsymbol{x_g})\bigr),
\end{equation}

where $D_E(i, j)$ is a distance metric (e.g., Euclidean distance as in our simulation). The Student $M{k}$, tries to minimize the distance $D_E$ in $n_s$ number of iteration where $n_s > n_g$.
\subsubsection{IL through KD loss} \label{KDloss}

To address catastrophic forgetting in IL, we utilize a data generator that employs input $x_g$ (initialized randomly) and the previously trained (now fixed) encoder $M_{k-1}$ to create synthetic images $\hat{D}_k$ representing classes from past tasks. The synthetic images $\hat{D}_k$ are fed into both $M_{k-1}$ and $M_k$. The model $M_{k-1}$ serves as the Teacher, transferring the knowledge of the previous tasks to the trainable Student $M_k$, which is learning current task $k$ simultaneously. This knowledge transfer is facilitated through the proposed KD loss, denoted as 
$\mathcal{L}_{\text{KD}}$:
\begin{equation}
\mathcal{L}_{\text{KD}} = \mathcal{L}_{\text{FAM}} + \mathcal{L}_{\text{Cov}} , \label{eqat}
\end{equation}
The first component $\mathcal{L}_{\text{FAM}}$ in equation (\ref{eqat}) corresponds to the feature attention matching loss, aimed at retaining information across all intermediate layers. Equation (4) represents an L2-norm-based feature alignment loss, which measures the difference in normalized feature maps between the Teacher and Student models across intermediate layers. This loss ensures that the Student retains important spatial and structural information from the Teacher, preventing significant feature drift. \( M_{k-1}(A_l) \) is the feature map at layer \( l \) from the Teacher model \( M_{k-1} \), and \( M_k(A_l) \) is the corresponding feature map at layer \( l \) from the Student model \( M_k \), which is being trained. \( A_l \) is the feature map at layer \( L \), and \( N_L \) is the total number of layers.

% \( M_{k-1}(A_l) \) refers to the feature map at layer \( l \) from the Teacher model \( M_{k-1} \), while \( M_k(A_l) \) refers to the corresponding feature map at layer \( l \) from the Student model \( M_k \), which is currently being trained.

\begin{equation}
\mathcal{L}_{\text{FAM}} = \sum\limits_{l=1}^{N_L} \bigg\Vert \frac{M_{k-1}(A_l)}{\parallel M_{k-1}(A_l)\parallel} - \frac{M_{k}(A_l)}{\parallel M_{k}(A_l)\parallel} \bigg\Vert_2,
\end{equation}

\noindent This summation encompasses all convolutional layers a part from the last average pooling layer, which is typically referred to as the embedding layer. The second term is covariance between the Teacher and the Student. We define the covariance for embedding matrix $Z$ (last layer in CNN) as:
\begin{equation}
C(Z)=\frac{1}{n-1}\sum_{i=1}^n(z_i-m){(z_i-m)}^t,\label{eq11}  
\end{equation}
$Z$ is the embedding matrix from the last layer of the CNN, $z_i$ is the embedding vector for the $i$-th sample, $n$ is the number of data samples in the batch, and $m$ is the mean of all embedding vectors. $C(Z)$ captures the relationship between feature dimensions in the Student and Teacher models and is used in $\mathcal{L}_{\text{Cov}}$ to regularize knowledge transfer.

\vspace{0.3cm} Inspired by Barlow \cite{Re36}, the covariance regularization term, c, can be defined as the adding of the squared-off diagonal coefficients of c(Z), with a factor 1/d, and d=512, which can cause the criterion to scale as a function of the dimension:
\begin{equation}
c(Z)=\frac{1}{d}\sum_{l\neq i}[C(Z)]_{l,i}^2,\label{eq12}        \end{equation}

then covariance loss between $M_k$ and $M_{k-1}$ is:
\begin{equation}
\mathcal{L}_{\text{Cov}} = C(Z_{k}) + C(Z_{k-1})
\end{equation} 
This term encourages the off-diagonal coefficients of $C(Z)$ between Teacher and Student to be close to 0, keeping previous information untouched. As we can see in Figure~\ref{fig:proposed-method}, $Z_k$ and $Z_{k-1}$ represent embeddings from the Student model $M_k$ (current task) and the Teacher model $M_{k-1}$ (previous tasks). These embeddings play a crucial role in multiple loss functions. In Triplet Loss $L_{\text{tri}}(x_k)$, $Z_k$ helps the Student model learn discriminative features for the current task. In Generated Loss $L_G(x_g)$, the difference between Student and Teacher embeddings guides the generator to create synthetic samples that resemble past data. In Covariance Loss, alignment between the covariance matrices of $Z_k$ and $Z_{k-1}$ ensures feature correlations remain stable across tasks. Putting all the parts together, the final training loss for the Student is as equation (\ref{eq14}).
% \vspace{-27pt}

% \vspace*{.17cm}

\begin{algorithm}[htbp]
\caption{Proposed Framework}
\scriptsize % or \footnotesize or \scriptsize
\begin{algorithmic}
\STATE Data: $D_k$ \\
\STATE Initialize: $G(\cdot;\phi)(VAE)$ \\
\STATE Teacher: $M_{k-1}$ \\
\STATE Student: $M_k$
\STATE \textbf{for} $k$ from $1$ to $K$
\STATE \quad \textbf{for} $1,2,\ldots,E$
\STATE \quad\quad $z \gets \mathcal{N}(0,I)$
\STATE \quad\quad \textbf{for} $1,2,\ldots,n_G$
\STATE \quad\quad\quad $x_g \gets G(z;\phi)$
\STATE \quad\quad\quad $\mathcal{L}_G \gets -D_E(M_k(x_g), M_{k-1}(x_g))$
\STATE \quad\quad\quad $\phi \gets \phi - \eta_g \cdot \nabla_{\phi} \mathcal{L}_G(\phi)$
\STATE \quad\quad \textbf{end}
\STATE \quad\quad \textbf{for} $1,2,\ldots,n_S$
\STATE \quad\quad\quad $\mathcal{L}_{M_k} \gets \mathcal{L}_{tri} + \lambda \mathcal{L}_{KD} + D_E(M_k(x_g), M_{k-1}(x_g))$
\STATE \quad\quad\quad $\theta_k \gets \theta_k - \eta_s \cdot \nabla_{\theta_k} \mathcal{L}_{M_k}(\theta_k)$
\STATE \quad\quad \textbf{end}
\STATE \quad \textbf{end}
\STATE \textbf{end}
\end{algorithmic}
\end{algorithm}

\vspace{-10pt} % Adjust this value as needed to control the spacing

% Force placement of the algorithm to the current location
\FloatBarrier 

% Add negative vertical space to reduce the gap
\vspace{5pt} % Adjust this value as necessary
% First table environment
\begin{table}[htbp]
\captionsetup{skip=10pt} % <-- ADD THIS before \caption
\centering
\begin{tabular}{lcc}
\hline
\textbf{Schemes} & \textbf{Average Accuracy (\%)} \\
\hline
& $K=2$ \\
\hline
Non-incremental learning (upper-bound) & 83.21 \\
Fine-tune (lower-bound) & 26.25 \\
\textbf{Our method} & 68.73 \\
\hline
\end{tabular}
\caption{Average accuracy for PI-CAI dataset.}
\label{tab:accuracy_picai}
\end{table}

\FloatBarrier 
% Add negative space to reduce the gap between the tables
\vspace{-10pt} % Adjust this value as needed to control the spacing
% \begin{table}[htbp]
% \centering
% \scriptsize % Smaller font size for the table
% \floatconts
%   {tblab}%
%   {\caption{Average accuracy, denoted as $A_K$ (\%), for ablation study of each term of our KD loss $\mathcal{L}_{\text{KD}}$.}}%
%   {\begin{tabular}{c c c}
%       \toprule
%       \textbf{Schemes} & \textbf{OCT} & \textbf{CIFAR-10} \\
%       \midrule
%       Fine-tune (lower-bound) & 33.33 & 32.20 \\
%       $\mathcal{L}_{\text{KD}}$ = $\mathcal{L}_{\text{FAM}}$ & 47.38 & 44.21 \\
%       $\mathcal{L}_{\text{KD}}$ = $\mathcal{L}_{\text{Cov}}$ & 49.65 & 46.14 \\
%       \textbf{Our method} & \textbf{64.43} & \textbf{67.23} \\
%       \bottomrule
%   \end{tabular}}
% \end{table}

\begin{table}[!t] % Force the table to be placed at the top of the page
\centering
% \scriptsize % Smaller font size for the table
% \floatconts
  
\vspace{7pt} 
  {\begin{tabular}{l c c c}
     \hline
      \textbf{Schemes} & \textbf{OCT} & \textbf{PathMNIST} & \textbf{CIFAR-10} \\
      \hline
      \textbf{Joint learning (upper-bound)} & 90.76 & 89.28 & 88.01 \\
      \textbf{Fine-tune (lower-bound)} & 33.33 & 28.89 & 32.20 \\
      LwF \cite{Re66} & 44.8 & 25.20 & 32.90 \\
      GR \cite{Re56} & 35.83 & 21.95 & 31.50 \\
      RWalk \cite{Re95} & 33.33 & 27.05 & 35.00 \\
      OWM \cite{Re94} & 38.93 & 52.42 & 48.30 \\
      EFT \cite{Re96} & 43.20 & \textbf{66.82} & 60.65 \\
      BIR \cite{Re97} & 62.00 & 35.17 & 64.68 \\
      \textbf{Our method} & \textbf{64.43} & 53.75 & \textbf{67.23} \\
   
  \end{tabular}}
\caption{Average accuracy, denoted as $A_K$ (\%), achieved upon completion of the final task. Our approach significantly surpasses other methods, including those based on coreset replay.}
\label{t27}
\end{table}

\section{Experiments}\label{se4}

\subsection{Experimental setting}\label{se4sub1}
\textbf{Dataset}: In our research, we used four different datasets to evaluate our method as: \textbf{PI-CAI} \cite{Re37}, optical coherence tomography \textbf{OCT Dataset} \cite{Re91}, \textbf{PathMNIST} \cite{Re92} and \textbf{CIFAR-10} \cite{Re93}. The first three datasets consist of medical images, while the fourth is CIFAR-10 (a general benchmark usually used in IL scenarios as well). 

\textbf{PI-CAI}: The Prostate Imaging Cancer AI (PI-CAI) dataset \cite{Re37} is used to investigate two distinct scenarios. For each scenario, we consider two tasks, each of which has two classes. First task contains {\it case-ISUP0} and {\it case-ISUP3}, while the second task includes {\it case-ISUP1} and {\it case-ISUP2}. To compare to other methods, we consider the same settings as in \cite{Re90} for the remaining three datasets: 

\textbf{OCT}: The OCT dataset \cite{Re91} includes more than 108,309 publicly accessible training images, categorized into four classes based on retinal conditions: Normal (47.21\%), Drusen (7.96\%), Choroidal Neovascularization (CNV) (34.35\%), and Diabetic Macular Edema (DME) (10.48\%). Each class has 250 corresponding testing images. In this dataset, we consider three tasks as in \cite{Re90}. The first task focuses on two classes: Normal and CNV, while the second task focuses solely on DME, and the final task addresses only Drusen. 

\textbf{PathMNIST}: PathMNIST \cite{Re92}, part of the MedMNIST collection, comprises histology slides representing nine distinct classes of colon pathology. The dataset contains 89,996 training and 7,180 testing images, with the number of training images per class ranging from 7,886 to 12,885, averaging around 10,000. These nine classes are organized into three tasks, each containing three classes as in \cite{Re90}.

\textbf{CIFAR-10}: The CIFAR-10 dataset \cite{Re93} serves as a conventional benchmark for continual learning, and consists of 60,000 natural images across 10 categories, including classes such as cats and trucks. Each class contains 5,000 training images and 1,000 testing images. The first task consists of four classes, while the next two tasks contain three classes, which is identical to \cite{Re90}.

\textbf{Implementation details}: As the trainable encoder, $M_k$, we use a ResNet-18 with the classifier and the fully-connected layer removed. Therefore, the embedding dimension $d$ is equal to 512. Adam is used as the optimizer for both generator and Student with the learning rate $\eta_g =0.001$ and $\eta_s = 0.00001$. The weight decay is $0.0001$ for both the generator and Student and the hyper-parameter $\lambda=0.8$. For each epoch $E$ we set $n_g=3$ and $n_s = 20$. We use a VAE generator with only three convolutional layers, the dimension of the input noise $z$ is 100. The mini-batch size $b$ for constructing triplet loss is $b=64$. The mini-batch size $n$ for constructing the KD losses is $n=16$. This means that, for constructing the total loss $L_{M_k}$ in each mini-batch, the total number of synthetic images is one-fourth of the current (new) actual images. We conducted the experiments using PyTorch on an NVIDIA RTX 3090 (24GB VRAM) for training.

\subsection{Comparison results}\label{se4sub2}
We employ the average accuracy to evaluate performance. Upon completing the training of all $K$ tasks, the average accuracy $A_K$ is determined by the formula: $A_K = \frac{1}{K} \sum_{j=1}^{K} a_{K,j}$. Here, $a_{i,j}$ denotes the test accuracy of the model, which has been incrementally trained from task 1 to task $i$ on the held-out test set of task $j$. In the first scenario, we compare our proposed method's average accuracy $A_K$, with fine-tuning as the lower bound, and non-incremental learning as the upper bound (only using the PI-CAI dataset). Table~\ref{tab:accuracy_picai} shows the average accuracy obtained from ten individual runs. We see that the accuracy of our proposed method is close to the upper bound and performs better than lower-band baseline. We assessed the average accuracy $A_K$ of our proposed approach in comparison to the baselines under the class-incremental learning (class-IL) scenario. Table~\ref{t27} shows the results for datasets divided identically to \cite{Re90} for a fair comparison. The results of our proposed method were obtained from averaging over ten separate trials, each using a different class order. Our approach consistently outperformed all baselines in OCT and CIFAR-10 datasets. Although \cite{Re90} achieves high accuracy for the PathMNIST dataset, the model’s dynamic nature and ability to grow network parameters to accommodate new tasks yields concerns about its associated Floating Point Operations (FLOP) requirements \cite{Re98}. Our KD loss, which keeps  covariance and feature maps between Teacher and Student remain the same during training, effectively combats catastrophic forgetting without using real data samples.
% \vspace{-20pt}
\subsection{Ablation Study}\label{se4sub3}
\begin{figure}[H]
\centering
\includegraphics[width=1\textwidth]{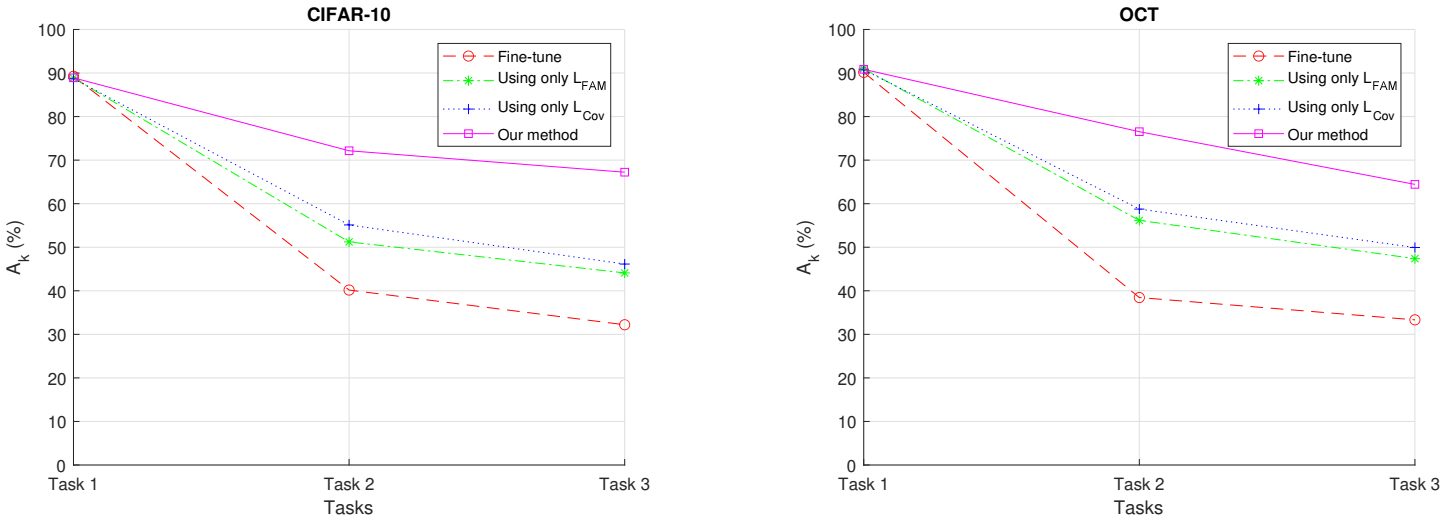}
\caption{$A_k$ (\%) for the ablation study conducted on two datasets.}
\label{fig:ablation}
\end{figure}

In this subsection, more experiments were conducted to reveal the effect of each term of our KD loss $\mathcal{L}_{\text{KD}}$ in~(\ref{eqat}). We present the results of ablation experiments to illustrate the effect of $\mathcal{L}_{\text{Cov}}$ and $\mathcal{L}_{\text{FAM}}$ separately. Table~\ref{tab:ablation_kd} shows the results of the ablation study which indicate that using feature attention matching loss $\mathcal{L}_{\text{FAM}}$ or covariance loss $\mathcal{L}_{\text{Cov}}$ individually leads to a significant reduction in accuracy across both datasets. Specifically, when employing feature attention matching loss alone, the average accuracy for OCT and CIFAR-10 is 47.38\% and 44.21\%, respectively. Similarly, utilizing covariance loss yields average accuracies of 49.65\% for OCT and 46.14\% for CIFAR-10. In contrast, our comprehensive method achieves a remarkable accuracy of 64.43\% for OCT and 67.23\% for CIFAR-10, underscoring the importance of integrating both loss components to optimize performance effectively. For a more detailed examination of the experiments conducted for each task, Figure~\ref{fig:ablation} presents the task-wise accuracy, which compares the performance of fine-tuning, using only feature attention matching loss $\mathcal{L}_{\text{FAM}}$, using only covariance loss $\mathcal{L}_{\text{Cov}}$, and applying both KD losses together, representing our full method. Figure~\ref{fig:ablation} clearly illustrates the variations in accuracy across each task, highlighting the effectiveness of our combined strategy in enhancing performance across the tasks.

% \section{Conclusion}\label{se4}
% In this paper we investigated class-incremental learning using Knowledge Distillation. Rather than storing past data samples, we used a recurrently updated encoder to generate synthetic images representing historical data. This approach helped construct KD regularizers to mitigate catastrophic forgetting in medical image datasets. Our results demonstrate the approach's efficacy, showing considerable reductions in catastrophic forgetting, and improved model performance across successive learning tasks.

\section{Conclusion}\label{se4}
In this paper, we investigated class-incremental learning using Knowledge Distillation (KD) to address the challenge of catastrophic forgetting in medical image analysis. Instead of storing data from previous tasks, we employed a recurrently updated encoder to generate synthetic images that approximate the distribution of historical data. These synthetic samples enabled the construction of effective KD-based regularization terms, allowing the model to retain prior knowledge while learning new tasks. Our results demonstrate the approach's efficacy, showing reductions in catastrophic forgetting, and improved model performance across successive learning tasks.
% In this paper we investigated class-incremental learning using Knowledge Distillation. Rather than storing past data samples, we used a recurrently updated encoder to generate synthetic images representing historical data. This approach helped construct KD regularizers to mitigate catastrophic forgetting in medical image datasets. Our results demonstrate the approach's efficacy, showing considerable reductions in catastrophic forgetting, and improved model performance across successive learning tasks.
% \clearpage
% \section*{Supplementary Figures and Tables}
\vspace{7pt} 
\begin{table}[htbp]
\centering
\begin{tabular}{lcc}
\hline
\textbf{Schemes} & \textbf{OCT} & \textbf{CIFAR-10} \\
\hline
Fine-tune (lower-bound) & 33.33 & 32.20 \\
$\mathcal{L}_{KD} = \mathcal{L}_{FAM}$ & 47.38 & 44.21 \\
$\mathcal{L}_{KD} = \mathcal{L}_{Cov}$ & 49.65 & 46.14 \\
\textbf{Our method} & \textbf{64.43} & \textbf{67.23} \\
\hline
\end{tabular}
\caption{Average accuracy, denoted as $A_K$ (\%), for ablation study of each term of our KD loss $\mathcal{L}_{KD}$.}
\label{tab:ablation_kd}
\end{table}

% \vspace{20pt}
\vspace{7pt} 

\clearpage

% % �� Now place the bibliography AFTER tables/figures
% \bibliographystyle{spiebib}
% \bibliography{report}

% \begin{table}[htbp]
% \centering
% \scriptsize % Smaller font size for the table
% \floatconts
%   {tblab}%
%   {\caption{Average accuracy, denoted as $A_K$ (\%), for ablation study of each term of our KD loss $\mathcal{L}_{\text{KD}}$.}}%
%   {\begin{tabular}{c c c}
%       \toprule
%       \textbf{Schemes} & \textbf{OCT} & \textbf{CIFAR-10} \\
%       \midrule
%       Fine-tune (lower-bound) & 33.33 & 32.20 \\
%       $\mathcal{L}_{\text{KD}}$ = $\mathcal{L}_{\text{FAM}}$ & 47.38 & 44.21 \\
%       $\mathcal{L}_{\text{KD}}$ = $\mathcal{L}_{\text{Cov}}$ & 49.65 & 46.14 \\
%       \textbf{Our method} & \textbf{64.43} & \textbf{67.23} \\
%       \bottomrule
%   \end{tabular}}
% \end{table}

% \begin{figure}[!t]
% \centering
% \includegraphics[width=1.1 \textwidth]{CIFAR OCT C-2.0.jpg}
% \caption{$A_k$ (\%) for the ablation study conducted on two datasets.}
% \label{fig4}
% \end{figure}

% \bibliographystyle{spiebib}
% \bibliography{report}
% References
% \bibliography{report} % bibliography data in report.bib
%\bibliographystyle{plain}
%\bibliography{report}

%\bibliographystyle{plain}
%\input{report.bbl}

% \bibliographystyle{spiebib} % makes bibtex use spiebib.bst
%\bibliography{midl-samplebibliography}
\clearpage
 
\textbf{Appendix A. Discriminative-separable Feature Space and Task Confusion:}\\

Task confusion is due to classifying different tasks that are not available together. It is a prime reason for degrading performance in a continual training regime. Using softmax loss in classification problems improved performance in many statics-learning scenarios. However, there is an inherent deﬁciency in softmax loss when it comes to IL. Softmax loss is mostly focused on producing linearly separable classes, as shown in  Figure~\ref{fig:fig2}(a). Softmax loss will cause the inter-class distance to be even smaller than the intra-class distance, which can be observed in  Figure~\ref{fig:fig2}(a), because softmax loss does not explicitly limit the ratio between the inter-class distance and intra-class distance. Therefore, in continual learning, classes in a new task will be confused with classes in the previous task due to the small distance between classes; making task confusion a severe problem of softmax loss.

One solution to the aforementioned problem is making the classes in embedding space separable, and as discriminative as possible. An example of a discriminative and separable case is illustrated in  Figure~\ref{fig:fig2}(b). One way to reach this goal is using generative models in which one model corresponds to learning each class \cite{Re34}, \cite{Re35}. However, the scalability of the generative method is a significant issue. Using deep metric learning with a similarity loss function is another practical solution to obtain both discrimination and separability. Triplet loss tries to put similar samples close to each other and produce large distances for dissimilar data points, which results in discriminative classes in the embedding space. Therefore, metric learning can be utilized as a powerful backbone in IL.

% Task Confusion results from the inability to effectively classify tasks presented separately, leading to performance degradation. While softmax loss performs well in static learning by achieving linear class separability, it fails in incremental learning (IL) due to its inability to maintain appropriate inter- and intra-class distances, and increases task confusion, impacting overall learning effectiveness.

% \vspace{20pt}

% \vspace{20pt} \textbf{Appendix B. Computational Efficiency in Image Generation using NVIDIA RTX 3090:}\\

% As shown in Table \ref{tab:computation_efficiency}, the computation efficiency, measured as the mean time per batch for generating a batch of 16 images over 100 iterations on the NVIDIA RTX 3090. This metric is useful to assess the speed of the  model during training.
% \setlength{\floatsep}{40pt}

% \begin{table}[ht]
% \centering
% \caption{Computation efficiency for generating a batch of 16 images with 100 iterations on NVIDIA RTX 3090}
% \label{tab:computation_efficiency}
% \begin{tabular}{@{}lc@{}}
% \toprule
% \textbf{Image Size} & \textbf{Mean Time per Batch (second)} \\ \midrule
% 32$\times$32$\times$3 & 1.27 \\ 
% 64$\times$64$\times$3 & 1.33 \\
% 84$\times$84$\times$3 & 1.45 \\
% 224$\times$224$\times$3 & 2.17 \\ \bottomrule
% \end{tabular}
% \end{table}
% ... (your last section, maybe Conclusion)

% \bibliographystyle{plain}
% \bibliography{report}

\end{document}